\documentclass[runningheads]{llncs}
\usepackage[T1]{fontenc}
\usepackage[caption=false,font=normalsize,labelfont=sf,textfont=sf]{subfig}
\usepackage{subfloat}
\usepackage{graphicx}
\usepackage{amsmath}
\usepackage{algorithm,algorithmic}
\usepackage{multirow} 
\usepackage{bigstrut}
\usepackage{color}
\usepackage{bm}
\usepackage{wrapfig}
\usepackage{pifont}
\usepackage{diagbox}
\usepackage{bbding}
\begin{document}
\title{
	Motion-Scenario Decoupling for Rat-Aware Video Position Prediction: Strategy and Benchmark
}

\author{Xiaofeng Liu\inst{1}
	\and
	Jiaxin Gao\inst{1}
	\and
	Yaohua Liu\inst{1}
	\and
	Nenggan Zheng\inst{2}
	\and  \\
	Risheng Liu\inst{1(}\Envelope\inst{)}
}
\authorrunning{Xiaofeng Liu et al.}
	\institute{Dalian University of Technology, Liaoning, China
	\\
	\email{rsliu@dlut.edu.cn}
	\\
	 \and
	Zhejiang University, Zhejiang, China
}

\maketitle              
\begin{abstract}
	Recently significant progress has been made in human action recognition and behavior prediction using deep learning techniques, leading to improved vision-based semantic understanding. However, there is still a lack of high-quality motion datasets for small bio-robotics, which presents more challenging scenarios for long-term movement prediction and behavior control based on third-person observation. In this study, we introduce RatPose, a bio-robot motion prediction dataset constructed by considering the influence factors of individuals and environments  based on predefined annotation rules. To enhance the robustness of motion prediction against these factors, we propose a Dual-stream Motion-Scenario Decoupling (\textit{DMSD}) framework that effectively separates scenario-oriented and motion-oriented features and designs a scenario contrast loss and motion clustering loss for overall training. With such distinctive architecture, the dual-branch feature flow information is interacted and compensated in a decomposition-then-fusion manner. Moreover, we demonstrate significant performance improvements of the proposed \textit{DMSD} framework on different difficulty-level tasks. We also implement long-term discretized trajectory prediction tasks to verify the generalization ability of the proposed dataset.

\end{abstract}

\section{Introduction}

In recent years, significant advancements in neuroscience and biomedical engineering have facilitated the development of Brain-Computer Interface (BCI)~\cite{wolpaw2000brain,nicolas2012brain,roy2022adaptive}, which offer potential benefits for rehabilitating neuromotor disorder~\cite{zhang2015control} and controlling small animals ~\cite{gupta2020hierarchical} such as rats, beetles and doves. These animals endowed with autonomous intelligence and dexterous and agile bodies are allowed to perform special military missions or emergency rescues after natural disasters where human intervention is difficult. However, the intention of animals and the uncertainty associated with changes in the environment and individual organisms pose significant challenges for effective manipulation by humans. Therefore, it is of great significance to acquire high-quality movement data and perform a comprehensive analysis of robotic movement behavior based on varying environmental and individual factors to regulate biological behavior.


The inference and prediction of human motion~\cite{moeslund2011visual,klette2008human} based on visual observation has been an extensively researched topic. Representative works include the HUMAN3.6M dataset~\cite{ionescu2013human3}, which predicts human pose, and the LMBRD dataset~\cite{jhuang2010automated}, which recognizes different behaviors of housed mice. Additionally, the 'something something' dataset~\cite{goyal2017something} is proposed for human-object interaction, which predicts actions performed by humans with respect to different objects. These datasets use videos to capture the meta-action and interaction between the target creatures and their surrounding environment, leading to improved predictions and better understanding of their behavior. However, these datasets generally focus on the current specific actions of organisms, without considering the movement outcomes in the future, which limits the exploration of the organisms' moving motivation.

Deep learning based action recognition methods~\cite{zolfaghari2018eco,zhou2018temporal} and large-scale video datasets have made significant progress in recent years, and different architectures or modules are proposed to improve the temporal modeling capabilities of deep learning models.
TRN~\cite{zhou2017temporalrelation} introduces a temporal relation network module to learn and reason about temporal dependencies between video frames at multiple time scales.
Similarly, TIN~\cite{shao2020temporal} improves the temporal feature extraction based on TSM and proposes a temporal interlacing operator to fuse the temporal and spatial information.
While these methods have shown some success in action recognition, their limitations in solely focusing on the temporal and spatial features between adjacent frames, without taking into account environmental and biological factors present in the video, thus hinder their performance in predicting the future movement of rats with high levels of uncertainty.

\subsection{Our Contribution}

To boost the investigation on motion behavior analysis of bio-robots and further research of behavioral control prototype incorporating animal moving intention, we first build a dataset RatPose which focuses on the bio-robot's movement prediction based on collected third-person video sequences. The movement data is collected in consideration of influence factors including varying individuals and different environments (e.g., Open Field and Maze), covering up to 5 scenarios, 6 individuals and 1023 data pairs. We also conduct detailed analysis of the individuals' differences and how it influences the performance of bio-robot's motion prediction. Based on the above analysis, we propose a Dual-stream Motion-Scenario Decoupling (\textit{DMSD}) framework to effectively decompose the scenario-oriented and motion-oriented features. Then we propose a motion clustering loss to measure the distance between features and the nearest simulated cluster center and combine the scenario contrast loss to construct the overall loss function. Finally, we demonstrate the significant performance improvement of the proposed \textit{DMSD} framework on single scenario and multiple scenarios motion prediction tasks. Furthermore, we implement our method to tackle more challenging discretized trajectory prediction based on long-term video sequences to show its generalization performance over different individuals and environments. In summary, our contributions can be summarized as follows:


\begin{itemize}
	
	\item Based on predefined annotation rules, we construct the first bio-robot's motion prediction dataset, namely RatPose, with full consideration of varying individuals and environments, and conduct comprehensive analysis of the influence of individuals and prediction intervals for proper task settings, which contributes to further research on the behavioral control prototype incorporating all kinds of uncertainties.
	\item We propose a Dual-stream Motion-Scenario Decoupling (\textit{DMSD}) framework to decompose the scenario-oriented and motion-oriented features, and design a novel motion clustering loss together with the scenario contrast loss to facilitate the robustness over varying environments and states of individuals.
	\item Extensive experiments under difference difficulty levels and ablation studies have demonstrate the effectiveness over existing methods ($34.3\%$ and $29.6\%$ relative improvement of Top 1 accuracy under single and multiple scenario, respectively). We also verify the generalization performance of this framework on more challenging discretized trajectory prediction tasks.
\end{itemize}

\section{Problem Set-up}
We focus on the long-term movements of rats to monitor their moving intentions, rather than their specific detailed actions, from an overhead perspective.
To approximately estimate the intention, we aim to predict the motion patterns of rats and model it into a classification problem, which has potential applications in various fields, including bio-robot controlling. 

\subsection{Basic Settings}
In this study, we address a unique problem that differs from the conventional action recognition tasks. While action recognition typically focuses on identifying what a human is doing or how they are interacting with objects, our goal is to predict the future moving position of rats. Unlike humans, animals do not always act logically or rationally, making this problem more challenging. Moreover, we use a different labeling strategy that is based on the outcome of the motion rather than the motion itself.

\begin{figure}[!h]
	\centering
	\includegraphics[width=0.95\linewidth]{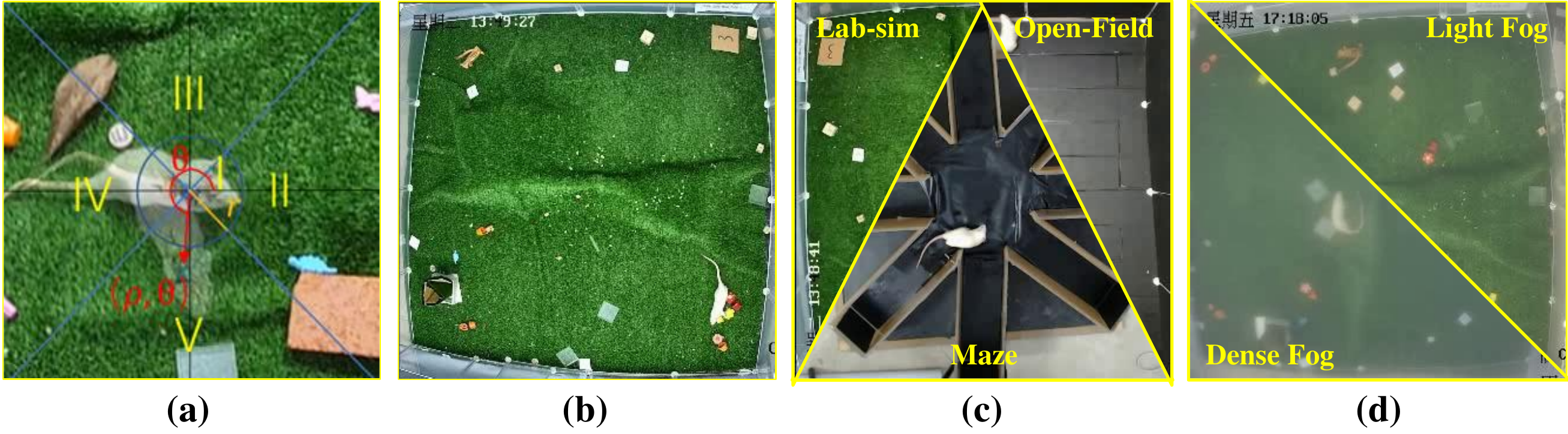}
		\caption{Problem definition (a) and multi-view task scenario configuration (b)-(d). Subfigure (a) illustrates the classification labels used under different tasks, while subfigures (b)-(d) illustrate three task settings based on the associated difficulty levels: single scenario, multiple scenarios, and more challenging scenarios. }    \label{tasks}
\end{figure} 

In our problem, we aim to classify motion results into five categories, namely top, down, left, right, and middle, as illustrated in Fig. \ref{tasks}. To determine the label of an object, we examine the distance it has moved relative to its body length. If the object moves a short distance, less than a predefined proportion of its body length, we label it as middle, indicating that it has stayed on the ground. Otherwise, we determine the label based on the direction of the body center's movement in pixel level. Specifically, we use $\theta$ to denote the counterclockwise angle between the width axis of the image and the motion vector after a short time $t$, $\rho$ to denote the pixel length of the motion vector, and $r$ to denote the predefined length in pixel level. Then we can define a set of the predicted motions: $\mathcal{M} = \{up:\rho > r , \frac{3}{4}\pi < \theta + 2k\pi< \frac{5}{4}\pi ; left: \rho > r, \frac{5}{4}\pi < \theta + 2k\pi< \frac{7}{4}\pi; down: \rho > r, \frac{7}{4}\pi < \theta + 2k\pi< \frac{9}{4}\pi; right: \rho > r, \frac{9}{4}\pi < \theta + 2k\pi< \frac{11}{4}\pi; middle: \rho < r \}$, where $k\in Z$.
Here we set $t=3s$ and $r$ is $1/10$ of the pixel length of the image.

Our problem can be characterized as a classification task with a unique focus and approach. In addition, our approach enables us to extract valuable insights from redundant video data in biological experiments, which may not be suitable for traditional action recognition tasks.

\subsection{Advanced settings}
Predicting the motion patterns of animals is a challenging task due to the unpredictable nature of their movements. A range of factors, including environmental, biological, and psychological aspects, contribute to the significant variance in animal motion outcomes. Therefore, it is crucial to consider these factors when developing a solution. In this study, we aim to address this issue by taking into account the visual differences between various scenarios and visual perception difficulties. To achieve this goal, we propose three different problem settings: a single scenario, multiple scenarios, and challenging scenarios ranging from easy to hard, as Fig. \ref{tasks} reveals.
The first problem setting involves collecting data from a few rats in a single scenario, while the second problem setting involves collecting data from a few rats in various scenarios. Finally, the third problem setting involves collecting data in a single scenario where the visual images are affected by simulated smoke using a mask placed on the camera. Except the visual challenge, there only exists 1 video per class for training or finetune in the challenging setting.
By incorporating these settings, we hope to develop a comprehensive understanding of animal motion patterns and contribute to the advancement of related research fields.

\section{The RatPose Dataset}
To better understanding rat's behaviors from visual perspective from a wide range of diverse environments and generalize to new settings,
we propose RatPose, a dataset for sharing different rats' moving experience in various scenarios.

\subsection{Data Collection Process}

\begin{figure}
    \centering
    \includegraphics[width=0.95\linewidth]{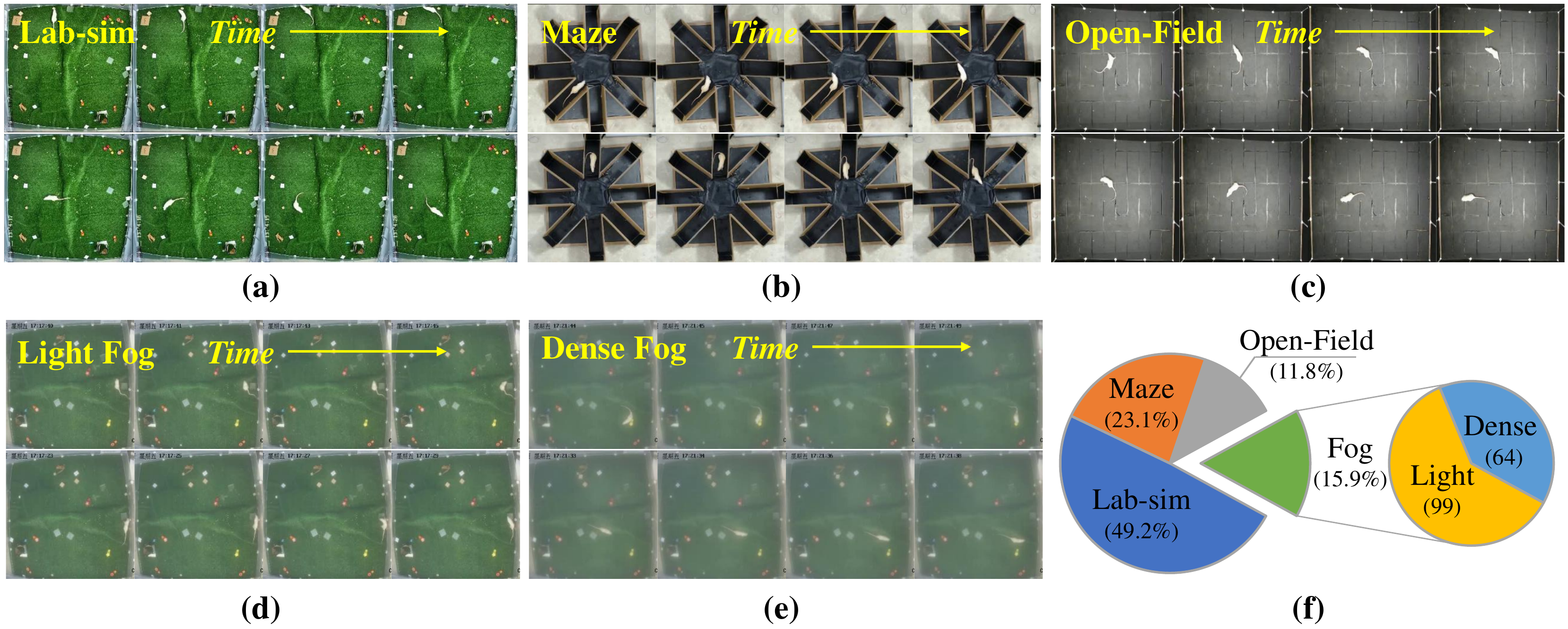}
    \caption{
        A glimpse of the RatPose video dataset. Scenario trajectories (a)-(e): Lab-sim, Maze, Open-Field, Light Fog and Dense Fog. (f): Data distribution. 
    }
    \label{visual data examples}
\end{figure}

Visual data was collected using an overhead camera to capture a vertical view of the environment, as illustrated in Fig. \ref{visual data examples}. To enhance the diversity of the data, videos were recorded of rats in various environments, including a simple maze with eight arms (Maze), a laboratory simulation of the real environment (Lab-sim), and an open field environment in a gym (Open Field). In order to expand the range of scenarios and increase the variability of the data, we considered the open-field environment as the baseline and introduced variations such as adding grass as the ground or placing toys as barriers.

\subsection{Dataset Features} 
\begin{figure}[!h]
	\centering
	\includegraphics[width=0.95\linewidth]{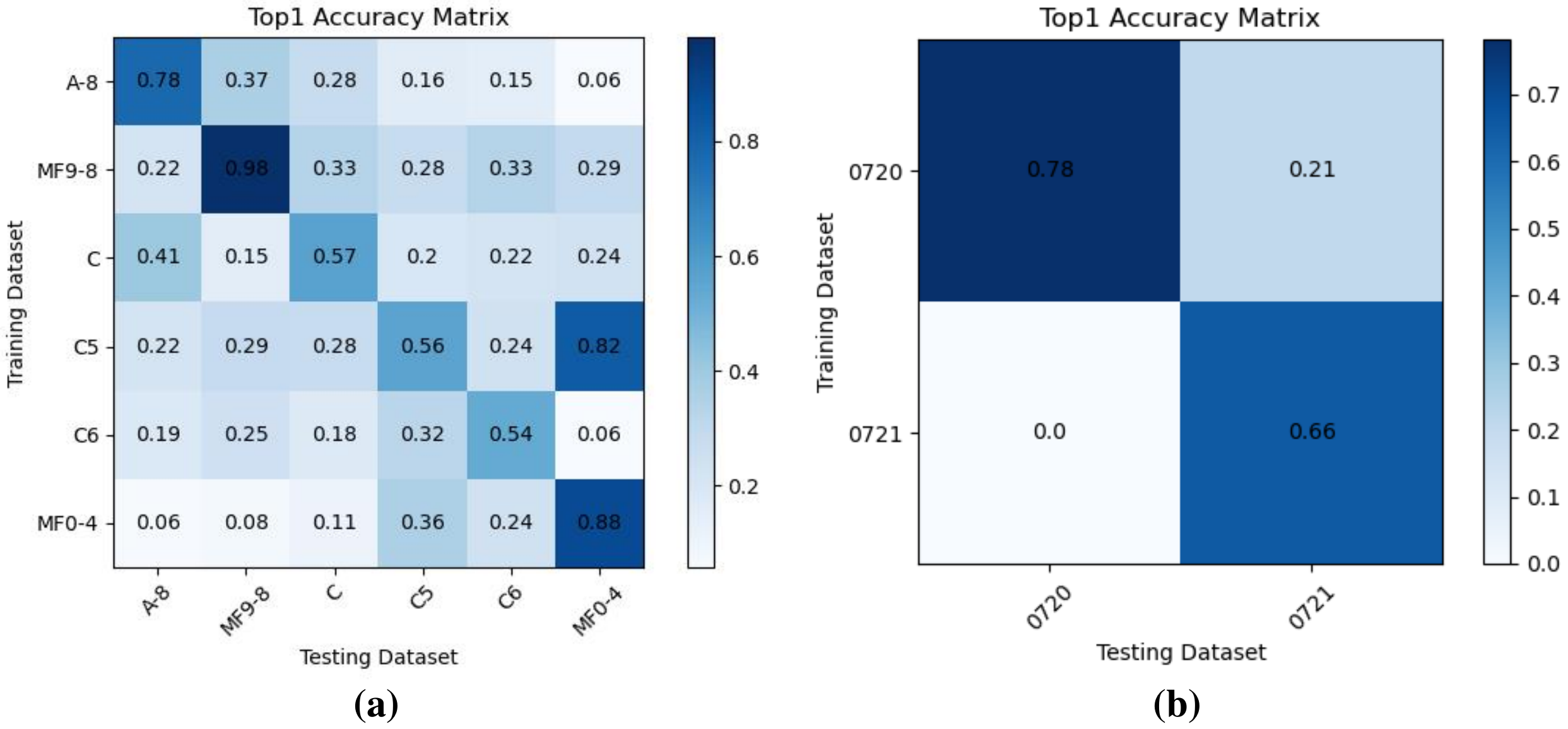}
		\caption{
		Performance using different data sets to train and test baseline model, each of which contains hundreds samples. (a): Testing performance using different rats ,named A-8, MF9-8, C, C5, C6 and MF0-4, to build data sets. (b): Testing performance using one rat, A-8, but in different time.
	}
	\label{meta_matrix_result}
\end{figure}

%

RatPose is full of challenging, and the challenging mainly comes from the natural insurance affected by many factors.
We primarily consider the influences caused by individual differences, and differences in individual states.
In order to demonstrate individual differences involved in modeling rat locomotor behavior, we utilize video data collected from six distinct rats.
As shown in \ref{meta_matrix_result}, 
the dataset is randomly partitioned into training and testing sets based on the individual rat, and a baseline model is trained using the training set and evaluated on multiple testing sets. 
Analysis of the resulting correlation matrix indicates that the model performs well in predicting the behavior of the corresponding rat, but struggles to generalize to new rats. 
And in the right subfigure, although the video data is collected with only a one-day time interval, there are notable variations in the state of the rat under observation. The observed heterogeneity in the rat's condition results in a substantial discrepancy between the predicted and actual outcomes of its behavioral actions.
These features suggest when facing the situation lack of sufficient data, model is fragile when meeting data full of diversity, especially in the situation that test data is not identically distributed.

\section{The Proposed Method}
\begin{figure}
	\centering
	\includegraphics[width=0.95\linewidth]{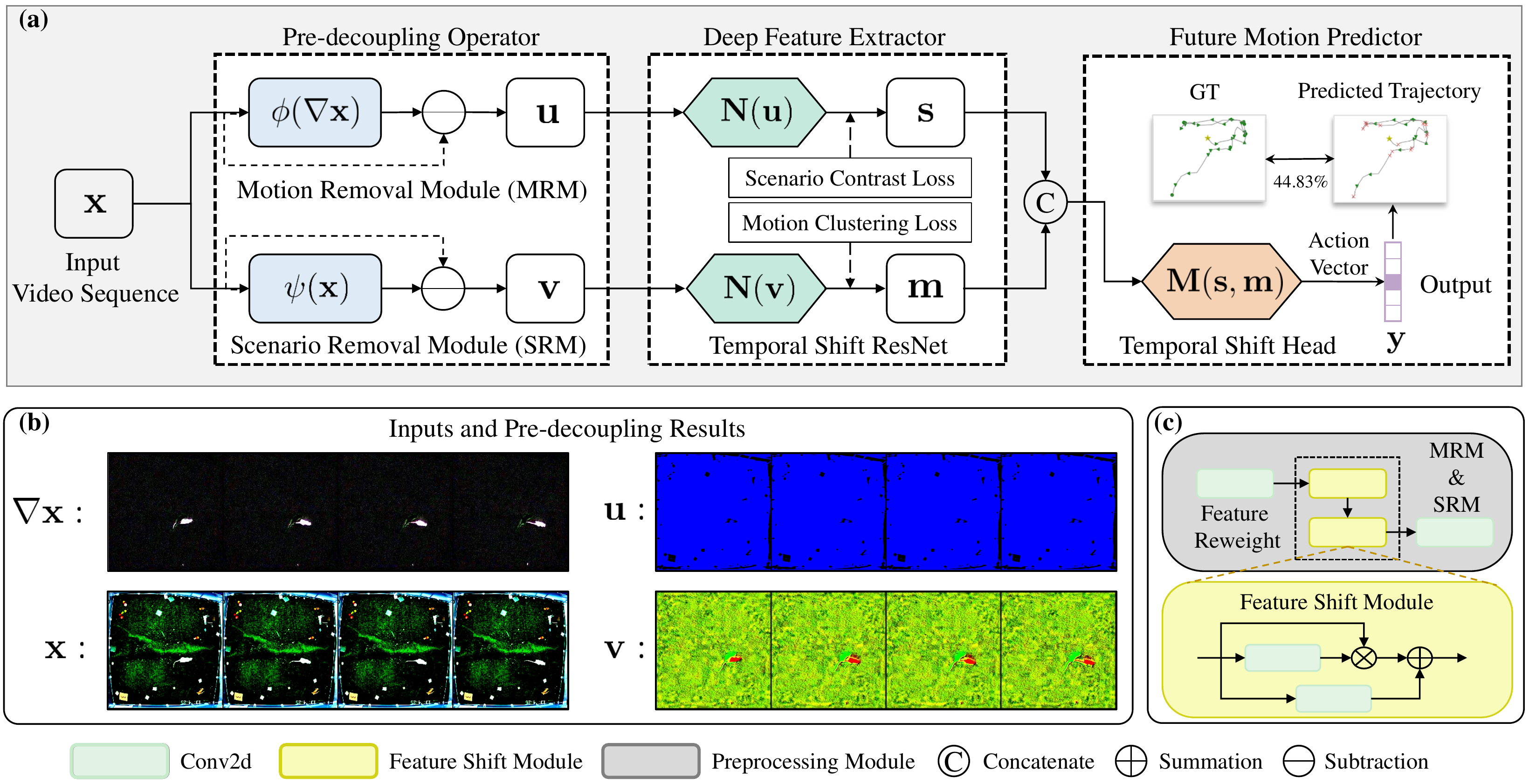}
	\caption{Overview of the proposed Dual-stream Motion-Scenario Decoupling (\textit{DMSD}) framework.  (a): Network architecture, including pre-decoupling operation, deep feature extractor and future motion predictor. (b): Visual images of the input and pre-decoupling results. (c): General network architecture of MRM and SRM. 
	}
	\label{network}
\end{figure}

This section describes the workflow of the Dual-stream Motion-Scenario Decoupling (\textit{DMSD}) framework and the training loss.  The overall architecture is shown in Fig. \ref{network}, mainly including three parts: the pre-decoupling operator, deep feature extractor, and the future motion predictor.  

\subsection{Dual-stream Motion-Scenario Decoupling (DMSD)}


As illustrated, pre-decoupling operator contains Motion Removal Module (MRM) denoted by $\phi$ and Scenario Removal Module (SRM) as $\psi$ to get the scenario relative input $\mathbf{u}$ and the motion relative input $\mathbf{v}$.
We then input the parallel pre-decoupling terms $\mathbf{u}$ and $\mathbf{v}$ into the dual feature extractor to obtain the deep scenario features $\mathbf{s}$ and the deep motion features $\mathbf{m}$ for prediction, with the guidance of scenario contrast loss and motion clustering loss.
The pre-decoupling operator and deep feature extractor compose a dual branch, jointly parameterized by $\theta$ and serves as the backbone to extract features. 
The future motion predictor uses temporal shift head~\footnote{See work~\cite{lin2019tsm} for the detailed implementation of temporal shift head.} denoted as $\mathbf{M}$ to fuse and output the action vector to predict the motion probability $\mathbf{y}$. The process can be formulated as 
\begin{equation}\label{eq: total}
	\mathbf{y} = M\big((N(\mathbf{x} - \phi(\mathbf{\nabla x})), N(\mathbf{x} - \psi(\mathbf{x})))\big),
\end{equation}
where  $\mathbf{x}=\{I_{t}| t=0,1,2,3...\}$ presents the input video sequence and $\mathbf{\nabla x} = \{I_{t} - I_0| t=0,1,2,3...\}$ presents the differences among $\mathbf{x}$, indicating movement of the target.

\textbf{Pre-decoupling Operator.}
To decompose the mixture components in a natural manner, we initially use MRM and SRM to decouple the input image sequences into different terms $\mathbf{v}$ and  $\mathbf{u}$.   
The MRM module is compromised by a convolution layer designed to expand the dimension, two feature shift module dedicated to recombining features and feature reweighting, and a convolution layer intended to reduce the dimension.
We apply MRM module $\phi$ on $\mathbf{\nabla x}$ and a residual-like subtraction operation to obtain the scenario relative term $\mathbf{u}$. 
We get this term by removing the motion component from the input sequences, enabling us to extract scenario features effectively.  
The SRM module shares the same network architecture with MRM while it has different parameters and input.
We apply SRM module $\psi$ directly on $\mathbf{x}$ to get a general representation of the input scenario.
Similarly, we use the subtraction operator as well as SRM to remove scenario information and obtain the motion relative term $\mathbf{v}$ by removing the background component from the original input. 
As shown in Figure \ref{network}, for input sequence $\mathbf{x}$, $\mathbf{u}$ present the scenario information and show the viable domain for rat's movement as the unreachable part is filled in black.
Meanwhile, $\mathbf{v}$ represents the movement of the rat in pixel space, with its spatial information.
By doing so, we pre-decouple the input sequence into different terms concentrated on different factors.

\textbf{Deep Feature Extractor.}
In addition to the pre-decoupling function, we use deep feature extractor for deep feature extraction and better factors decoupling.
The deep feature extractor  the deep feature extractor contains two temporal shift resnet~\footnote{See work~\cite{lin2019tsm,he2016deep} for the detailed implementation of temporal shift resnet.} denoted as $\mathbf{N}$ to extract and fuse the temporal and spatial features of $\mathbf{u}$ and $\mathbf{v}$.
Besides, we propose the scenario contrast loss and motion clustering loss \footnote{We provide more details about the proposed scenario contrast loss and motion clustering loss in section \ref{loss}.} to better detect the correlations among data and decouple the coupling relationship between $\mathbf{s}$ and $\mathbf{m}$. 

\textbf{Future Motion Predictor.}
After applying the dual branch, we utilize the Future Motion Predictor to fuse $\mathbf{s}$ and $\mathbf{m}$ features and generate the predicted output $\mathbf{y}$. Future Motion Predictor facilitates the integration and decoding of temporal and spatial features, alongside motion and scenario features, to yield robust and effective prediction outcomes. Notably, our approach demonstrate remarkable resilience to changes in environmental perspectives, with a prediction accuracy of up to $48.8\%$ even under such conditions.

\subsection{Loss Function} \label{loss}
We propose several losses to explore the similarity of action behaviors in different scenarios and the differences in scenario factors under different states. Among them, we optimize both dual-branch network's parameters $\theta$ and the clustering centers $\mathbf{r}$ using the feature decoupling loss $L_{f}$, which is expressed as follows:
\begin{equation}\label{aux_loss}
	L_{f}=\lambda_s \cdot L_{sc} + \lambda_m \cdot L_{mc}, 
\end{equation}
where $L_{sc}$ and $L_{mc}$ are the proposed scenario contrast loss and motion clustering loss. $\lambda_s$ and $\lambda_m$ are the corresponding weights. In our training, we empirically set $\lambda_s=0.1$ and $\lambda_m=1$.
In addition, we introduce the classification loss $L_{cls}$ optimize the whole networks parameters $\theta$ and $\omega$. The process of optimization is carried out alternately in every iteration step.

\textbf{Scenario Contrast Loss:}
In order to increase the sensitivity of the scene feature representation to slight changes in the environment, we utilize the technique of contrastive learning. We consider videos captured in the same scenario at the same time as positive samples, denoted as $\bm{s}^+$, while all others are considered as negative samples, denoted as $\bm{s}^-$. Drawing inspiration from contrastive learning for visual representations~\cite{chen2020simple}, we formulate scenario contrast loss as:

\begin{equation}\label{scenario_contrast_loss}
	L_{sc} = -\log\frac{exp(sim(\bm{s}, \bm{s}^+))}{\sum_{\bm{s}'\in \bm{s}^-} exp(sim(\bm{s}, \bm{s}'))}.
\end{equation}

\textbf{Motion Clustering Loss:}
We employ a series of trainable representation clustering centers $\mathbf{r}_c$ with label $c$. To measure the distance between each feature and the nearest cluster center, we calculate the minimum $L2$ distance between the feature and the center as the nearest auxiliary distance:
\begin{equation}\label{distance}
	D(\mathbf{m} | c) = \min_\mathbf{r} ||\mathbf{m} - \mathbf{r}||_2 \quad s.t. \quad \mathbf{r} \in \mathbf{r}_c.
\end{equation}
Next, we use the softmax function on the opposite number of the nearest auxiliary distance between each class feature center as the probability and apply cross-entropy to obtain the final motion clustering loss. This approach allows us to learn distributions of feature representations and obtain better decision boundaries:
\begin{equation}\label{Motion Clustering Loss}
	L_{mc} = CE(\frac{e^{-D(\mathbf{m}|\mathbf{z})}}{\sum_{k} e^{-D(\mathbf{m}|\mathbf{z})}}, \mathbf{z}).
\end{equation}
Here, $\mathbf{z}$ denotes the ground truth label, and $CE$ denotes the cross-entropy loss.

\section{Experiments}
In this study, all models are trained and tested on a server equipped with an Intel(R) Xeon(R) Gold 5218 CPU @ 2.30GHz and an NVIDIA A40 GPU. The training process is conducted using the mmaction framework~\cite{2020mmaction2} with default training hyper-parameters. We adopt a sampling strategy where every 8th frame of the video is selected and resized to $224\times224$. For each video, we sample the last 8 frames according to this rule. For longer video online prediction, we predict the position distribution every 3 seconds for the subsequent 3 seconds.
\subsection{Quantitative Evaluation}

\begin{table}[htbp]
	\centering
	\caption{
		Quantitative comparison of rat location prediction accuracy for state-of-the-art methods in Single and Multiple Scenarios.
	}
	\renewcommand\arraystretch{1.2}
	\setlength{\tabcolsep}{1.2mm}{
		\begin{tabular}{|c|c|c|c|c|c|c|c|c|c|c|c|}
			\hline
			\multicolumn{2}{|c|}{\multirow{2}{*}{\diagbox[width=5.1em,height=2.85em]{Metric}{Setting}}} & \multicolumn{5}{c|}{\textit{Single}} & \multicolumn{5}{c|}{\textit{Multiple}} \\
			\cline{3-12}\multicolumn{2}{|c|}{} & TRN & TSM & TIN & TSF & Ours & TRN & TSM & TIN & TSF & Ours \\
			\hline
			\multirow{2}[2]{*}{Acc} &  Mean$\uparrow$ & 37.39  & 37.39  & 39.92  & 33.68  & \textcolor[rgb]{ 1,  0,  0}{44.66} & 40.45  & 40.00  & 36.58  & 38.40  & \textcolor[rgb]{ 1,  0,  0}{46.68} \\
			& Std$\downarrow$ & 49.63  & 49.63  & 30.49  & 35.18  & \textcolor[rgb]{ 1,  0,  0}{5.01} & 37.39  & 42.23  & 28.21  & 27.66  & \textcolor[rgb]{ 1,  0,  0}{10.71} \\
			\hline
			\multicolumn{2}{|c|}{Top1Acc$\uparrow$} & 30.10  & 30.10  & 33.98  & 29.13  & \textcolor[rgb]{ 1,  0,  0}{45.63} & 34.98  & 34.98  & 32.51  & 34.48  & \textcolor[rgb]{ 1,  0,  0}{45.32} \\
			\hline
		\end{tabular}%
	}
	\label{tab:results}%
\end{table}%

In order to evaluate the effectiveness of our proposed method, we conducted experiments on both single and multiple scenarios tasks, as table \ref{tab:results} reveals. To demonstrate the superiority of our approach, we performed a quantitative comparison with four state-of-the-art methods, namely TRN~\cite{zhou2017temporalrelation}, TSM~\cite{lin2019tsm}, TIN~\cite{shao2020temporal}, and TSF~\cite{bertasius2021space}. The experimental results clearly indicate that our method achieves significantly better performance than the other methods in terms of both mean class accuracy and top-1-accuracy, providing strong evidence for the effectiveness of our proposed approach.

\begin{wrapfigure}{r}{6cm}
	\vspace{-0.2cm} %
	\begin{minipage}[t]{1.0\linewidth}
		\centering
		\makeatletter\def\@captype{table}\makeatother\caption{Ablation analysis of different loss functions in our proposed \textit{DMSD} training: Impact of Scenario Contrast Loss and Motion Clustering Loss.}\label{tab:ablation}
		\renewcommand\arraystretch{1.2}
		\setlength{\tabcolsep}{1mm}{
			\begin{tabular}{|c|c|c|c|c|c|}
				\hline
				&$L_{cls}$ &$L_{sc}$ &$L_{mc}$ & $Acc_{s}$$\uparrow$ & $Acc_{m}$$\uparrow$ \\
				\hline
				\footnotesize$S_1$ & \footnotesize \ding{52}  &\footnotesize \ding{56} &\footnotesize \ding{56} &\footnotesize $30.10$ &\footnotesize $42.36$\\
				\hline
				\footnotesize$S_2$ & \footnotesize \ding{52}  &\footnotesize \ding{52} &\footnotesize \ding{56} &\footnotesize $38.83$ &\footnotesize $41.38$ \\
				\hline
				\footnotesize$S_3$ & \footnotesize \ding{52}  &\footnotesize \ding{56} &\footnotesize \ding{52} &\footnotesize $39.81$ &\footnotesize $44.83$ \\
				\hline
				Ours &\footnotesize \ding{52}  &\footnotesize \ding{52} &\footnotesize \ding{52} &\footnotesize \textcolor[rgb]{ 1,  0,  0}{$46.68$} &\footnotesize \textcolor[rgb]{ 1,  0,  0}{$45.32$} \\ 
				\hline
			\end{tabular}
		} 
	\end{minipage}
	\vspace{-0.4cm}
\end{wrapfigure}

We conduct ablation experiments on our auxiliary loss, as table \ref{tab:ablation} reveals.
The $S_1$ approach represents the network trained solely with the classification loss, while $S_2$ and $S_3$ show the results obtained by integrating the Scenario Contrast and the Motion Clustering Loss, respectively. $Acc_{s}$ and $Acc_{m}$ represent the top-1 accuracy in single scenario and multiple scenarios respectively. To train the approaches other than $S_1$ an iterative joint training strategy was applied.

To demonstrate the efficacy of our training strategy, we conducted a comprehensive comparative analysis against several state-of-the-art few shot learning techniques, including RHG~\cite{pmlr-v70-franceschi17a}, BDA~\cite{BDA}, and IAPTT~\cite{liu2023augmenting}. In this context, we regarded the feature extraction layer as an upper-level challenge while considering the subsequent classification layer as a lower-level problem. Employing a bilevel optimization approach facilitated the seamless acquisition of a robust universal feature extraction module.
As evident from the results presented in Table~\ref{tab:bilevel}, traditional methods for few-shot learning struggle to acquire a viable representation and, in some cases, even perform worse than direct training. In stark contrast, our proposed approach not only surpasses the performance of other methods but also significantly enhances the final outcome. The superiority of our technique in capturing essential features and its ability to adapt to limited training data highlight its potential as a groundbreaking solution in the field of few-shot learning.

\begin{table}[htbp]
	\centering
	\caption{
		Quantitative comparison of rat location prediction accuracy for state-of-the-art methods with Different Training Strategies.
	}
	\renewcommand\arraystretch{1.2}
	\setlength{\tabcolsep}{1.2mm}{
		\begin{tabular}{|c|c|c|c|c|c|c|c|c|c|}
			\hline
			{\multirow{2}{*}{\diagbox[width=5.1em,height=2.85em]{Metric}{Setting}}} & \multicolumn{4}{c|}{\textit{TIN}} & \multicolumn{5}{c|}{\textit{DMSD}} \\
			\cline{2-10}\multicolumn{1}{|c|}{}& RHG & BDA & IAPTT & -  & RHG & BDA & IAPTT & - & Ours \\
			\hline
			 MeanAcc$\uparrow$ & 27.53  & 28.08  & 34.55  & 39.92  & 23.47 & 27.98  & 28.03  & 42.36 & \textcolor[rgb]{ 1,  0,  0}{46.68}   \\
			\hline
			{Top1Acc$\uparrow$} & 26.11  & 28.57  & 29.06  & 33.98  & 25.12  & 26.60  & 27.59 & 30.10 & \textcolor[rgb]{ 1,  0,  0}{45.32} \\
			\hline
		\end{tabular}%
	}
	\label{tab:bilevel}%
\end{table}%

We observe that utilizing the motion clustering loss and scenario contrast loss resulted in significant improvements in one or more of the evaluation metrics presented in the table. Furthermore, the combined use of these two losses during training leads to a substantial increase in the performance of the final model, particularly in a single scenario setting, where the top-1 accuracy is improved by approximately $52\%$ compared to the basic dual-branch model. This indicates that the motion clustering loss and scenario contrast loss are effective in improving the overall performance of the model.

\subsection{Qualitative Evaluation}
\begin{figure}[!h]
    \centering
    \includegraphics[width=0.95\linewidth]{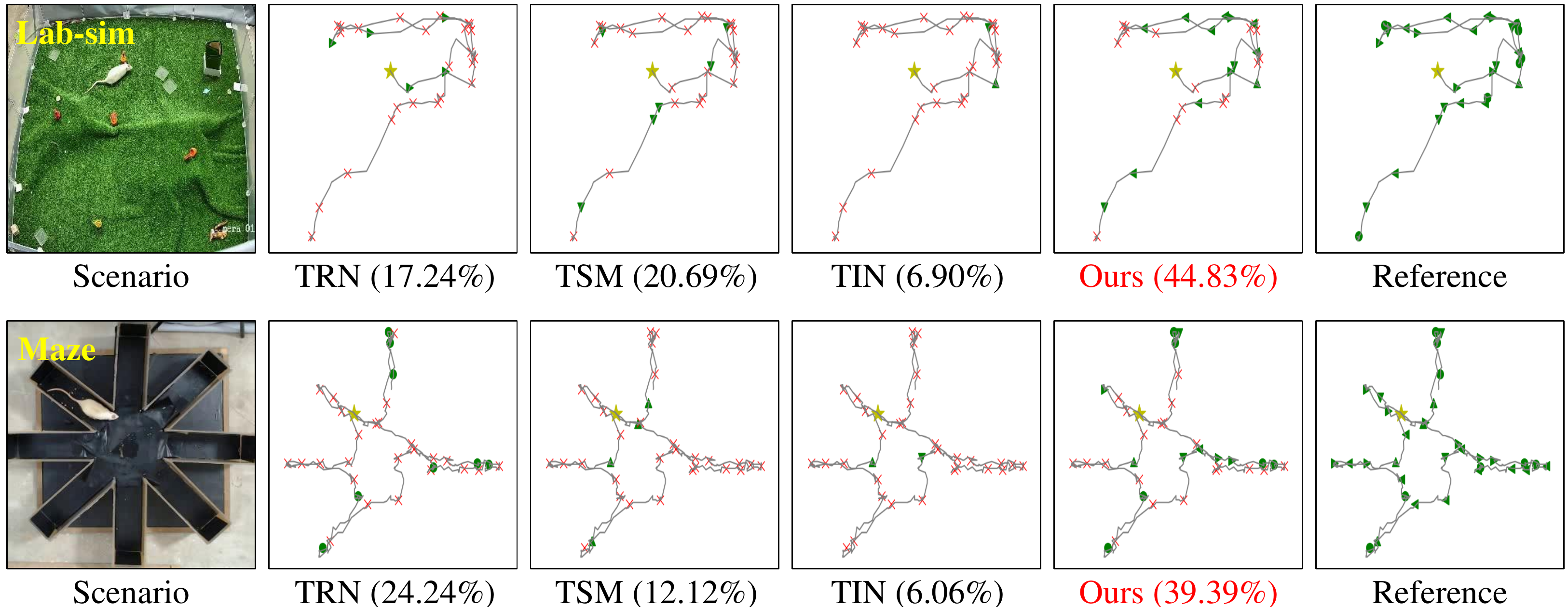}
    \caption{
        Qualitative comparison for tracking and predicting the movement trajectory of rats in different camera scenario views.
        Gray lines depict the trajectory of rat movement, circles indicate stationary positions, while triangles in up, down, left, and right directions represent the rat's movement.
        The red crosses signify prediction errors, and the yellow star pattern denotes the rat's starting position.
        The numerical value represents the top-1 accuracy on this trajectory.
    }
    \label{trajectories}
\end{figure}

To illustrate the generalizability and robustness of our method, we conduct qualitative evaluation employ several highly challenging video segments for practical experimentation. Notably, the data in these segments differs not only in the individual rats used for data collection but also in the camera angles employed for filming, as compared to those present in the RatPose training set.

As depicted in Fig. \ref{trajectories}, the first row of the figure shows a difference between the shooting angle of the video footage and the angle of the training data. Despite this slight visual discrepancy, the prediction accuracy of TRN and other methods considerably decline. In contrast, our method is found to be robust to changes in shooting angles, as its prediction accuracy remained unaffected.

Additionally, the rats in the second row of the figure display more vigorous movements than those in the dataset, performing complex activities within a short period in the maze environment. This leads to a decrease in prediction accuracy for all methods, reflecting the notable individual differences among rats and the difficulty of real-time prediction of their movement intentions. Nonetheless, our method was able to maintain a relatively reasonable level of prediction accuracy even under these challenging circumstances, further attesting to the superiority of our approach.

\section{Conclusion}
In conclusion, recent advancements in bio-robotics with Brain-Computer Interface (BCI) techniques have made it possible to directly control animals such as rats, beetles, and doves by humans. However, the highly unpredictable and difficult to control behaviors of these animals make it crucial to develop accurate and effective methods for predicting their movements. This paper proposes a novel dataset called RatPose for predicting the heading direction of rats' movement, which shifts the attention towards forecasting the future movement outcomes of organisms. The paper also proposes a visual prediction model that effectively decouples motion from the environment for this dataset, achieving superior performance when compared to state-of-the-art action recognition methods. 

\subsubsection{Acknowledgements} 		This work is partially supported by the National Key R\&D
Program of China (2020YFB1313503), the National Natural
Science Foundation of China (U22B2052), the Fundamental
Research Funds for the Central Universities and the Major
Key Project of PCL (PCL2021A12).

\bibliographystyle{unsrt}
\bibliography{bibfile}

\begin{thebibliography}{10}

\bibitem{wolpaw2000brain}
Jonathan~R Wolpaw, Niels Birbaumer, William~J Heetderks, Dennis~J McFarland,
  P~Hunter Peckham, Gerwin Schalk, Emanuel Donchin, Louis~A Quatrano, Charles~J
  Robinson, Theresa~M Vaughan, et~al.
\newblock Brain-computer interface technology: a review of the first
  international meeting.
\newblock {\em IEEE transactions on rehabilitation engineering}, 8(2):164--173,
  2000.

\bibitem{nicolas2012brain}
Luis~Fernando Nicolas-Alonso and Jaime Gomez-Gil.
\newblock Brain computer interfaces, a review.
\newblock {\em sensors}, 12(2):1211--1279, 2012.

\bibitem{roy2022adaptive}
Arunabha~M Roy.
\newblock Adaptive transfer learning-based multiscale feature fused deep
  convolutional neural network for eeg mi multiclassification in
  brain--computer interface.
\newblock {\em Engineering Applications of Artificial Intelligence},
  116:105347, 2022.

\bibitem{zhang2015control}
Rui Zhang, Yuanqing Li, Yongyong Yan, Hao Zhang, Shaoyu Wu, Tianyou Yu, and
  Zhenghui Gu.
\newblock Control of a wheelchair in an indoor environment based on a
  brain--computer interface and automated navigation.
\newblock {\em IEEE transactions on neural systems and rehabilitation
  engineering}, 24(1):128--139, 2015.

\bibitem{gupta2020hierarchical}
Akshansh Gupta, RK~Agrawal, Jyoti~Singh Kirar, Baljeet Kaur, Weiping Ding,
  Chin-Teng Lin, Javier Andreu-Perez, and Mukesh Prasad.
\newblock A hierarchical meta-model for multi-class mental task based
  brain-computer interfaces.
\newblock {\em Neurocomputing}, 389:207--217, 2020.

\bibitem{moeslund2011visual}
Thomas~B Moeslund, Adrian Hilton, Volker Kr{\"u}ger, and Leonid Sigal.
\newblock {\em Visual analysis of humans}.
\newblock Springer, 2011.

\bibitem{klette2008human}
Reinhard Klette, Dimitris~N Metaxas, and Bodo Rosenhahn.
\newblock {\em Human Motion: Understanding, Modelling, Capture, and Animation}.
\newblock Springer, 2008.

\bibitem{ionescu2013human3}
Catalin Ionescu, Dragos Papava, Vlad Olaru, and Cristian Sminchisescu.
\newblock Human3. 6m: Large scale datasets and predictive methods for 3d human
  sensing in natural environments.
\newblock {\em IEEE transactions on pattern analysis and machine intelligence},
  36(7):1325--1339, 2013.

\bibitem{jhuang2010automated}
Hueihan Jhuang, Estibaliz Garrote, Xinlin Yu, Vinita Khilnani, Tomaso Poggio,
  Andrew~D Steele, and Thomas Serre.
\newblock Automated home-cage behavioural phenotyping of mice.
\newblock {\em Nature communications}, 1(1):1--10, 2010.

\bibitem{goyal2017something}
Raghav Goyal, Samira Ebrahimi~Kahou, Vincent Michalski, Joanna Materzynska,
  Susanne Westphal, Heuna Kim, Valentin Haenel, Ingo Fruend, Peter Yianilos,
  Moritz Mueller-Freitag, et~al.
\newblock The" something something" video database for learning and evaluating
  visual common sense.
\newblock In {\em Proceedings of the IEEE international conference on computer
  vision}, pages 5842--5850, 2017.

\bibitem{zolfaghari2018eco}
Mohammadreza Zolfaghari, Kamaljeet Singh, and Thomas Brox.
\newblock Eco: Efficient convolutional network for online video understanding.
\newblock In {\em Proceedings of the European conference on computer vision
  (ECCV)}, pages 695--712, 2018.

\bibitem{zhou2018temporal}
Bolei Zhou, Alex Andonian, Aude Oliva, and Antonio Torralba.
\newblock Temporal relational reasoning in videos.
\newblock In {\em Proceedings of the European conference on computer vision
  (ECCV)}, pages 803--818, 2018.

\bibitem{zhou2017temporalrelation}
Bolei Zhou, Alex Andonian, Aude Oliva, and Antonio Torralba.
\newblock Temporal relational reasoning in videos.
\newblock {\em European Conference on Computer Vision}, 2018.

\bibitem{shao2020temporal}
Hao Shao, Shengju Qian, and Yu~Liu.
\newblock Temporal interlacing network.
\newblock {\em AAAI}, 2020.

\bibitem{lin2019tsm}
Ji~Lin, Chuang Gan, and Song Han.
\newblock Tsm: Temporal shift module for efficient video understanding.
\newblock In {\em Proceedings of the IEEE International Conference on Computer
  Vision}, 2019.

\bibitem{he2016deep}
Kaiming He, Xiangyu Zhang, Shaoqing Ren, and Jian Sun.
\newblock Deep residual learning for image recognition.
\newblock In {\em Proceedings of the IEEE conference on computer vision and
  pattern recognition}, pages 770--778, 2016.

\bibitem{chen2020simple}
Ting Chen, Simon Kornblith, Mohammad Norouzi, and Geoffrey Hinton.
\newblock A simple framework for contrastive learning of visual
  representations.
\newblock In {\em International conference on machine learning}, pages
  1597--1607. PMLR, 2020.

\bibitem{2020mmaction2}
MMAction2 Contributors.
\newblock Openmmlab's next generation video understanding toolbox and
  benchmark.
\newblock \url{https://github.com/open-mmlab/mmaction2}, 2020.

\bibitem{bertasius2021space}
Gedas Bertasius, Heng Wang, and Lorenzo Torresani.
\newblock Is space-time attention all you need for video understanding?
\newblock In {\em ICML}, volume~2, page~4, 2021.

\bibitem{pmlr-v70-franceschi17a}
Luca Franceschi, Michele Donini, Paolo Frasconi, and Massimiliano Pontil.
\newblock Forward and reverse gradient-based hyperparameter optimization.
\newblock In Doina Precup and Yee~Whye Teh, editors, {\em Proceedings of the
  34th International Conference on Machine Learning}, volume~70 of {\em
  Proceedings of Machine Learning Research}, pages 1165--1173. PMLR, 06--11 Aug
  2017.

\bibitem{BDA}
Risheng Liu, Pan Mu, Xiaoming Yuan, Shangzhi Zeng, and Jin Zhang.
\newblock A general descent aggregation framework for gradient-based bi-level
  optimization.
\newblock {\em IEEE Transactions on Pattern Analysis and Machine Intelligence},
  45(1):38--57, 2023.

\bibitem{liu2023augmenting}
Risheng Liu, Yaohua Liu, Shangzhi Zeng, and Jin Zhang.
\newblock Augmenting iterative trajectory for bilevel optimization:
  Methodology, analysis and extensions, 2023.

\end{thebibliography}
\end{document}